# A New 3D Method to Segment the Lumbar Vertebral Bodies and to Determine Bone Mineral Density and Geometry

André Mastmeyer, Klaus Engelke, Sebastian Meller, Willi A Kalender

Institute of Medical Physics, Henkestr. 91,
91052 Erlangen, Germany
{andre.mastmeyer, klaus.engelke}@imp.uni-erlangen.de
http://www.imp.uni-erlangen.de

**Abstract.** In this paper we present a new 3D segmentation approach for the vertebrae of the lower thoracic and the lumbar spine in spiral computed tomography datasets. We implemented a multi-step procedure. Its main components are deformable models, volume growing, and morphological operations. The performance analysis that included an evaluation of accuracy using the European Spine Phantom, and of intra-operator precision using clinical CT datasets from 10 patients highlight the potential for clinical use. The intra-operator precision of the segmentation procedure was better than 1% for Bone Mineral Density (BMD) and better than 1.8% for volume. The long-term goal of this work is to enable better fracture prediction and improved patient monitoring in the field of osteoporosis. A true 3D segmentation also enables an accurate measurement of geometrical parameters that can augment the classical measurement of BMD.

## 1 Introduction

**Aim.** Low bone mineral density is one of the most important risk factors for osteoporosis or to be more precise for spine and hip fractures. Traditionally conventional single slice Quantitative Computed Tomography (QCT) has been one of the methods used for the assessment of BMD in the spine. Specifically, single mid-vertebral slices of the three vertebral bodies L1-L3 are acquired and analyzed quantitatively for BMD [1] (see Fig. 1). Compared to other methods in the field one of the disadvantages of QCT was its higher precision error. The advent of multi slice spiral CT technology combined with advanced methods for the reduction of radiation exposure now offers the possibility for true volumetric tomographic acquisition protocols and analysis procedures that should improve precision. Along with the other advantages of QCT such as the separation of cortical and trabecular bone and the possibility to obtain volumetric geometrical measurements the improved precision should significantly amend the diagnostic value of QCT. However, advanced 3D segmentation methods are required in order to automatically define volumes of interest (VOIs) for which BMD will be determined. The aim of this contribution is the development and characterization of a 3D segmentation procedure for vertebral bodies.

**State of the Art.** Lang et. al. [2] showed improved precision of BMD measurements when using volumetric CT. However, they did not apply a full 3D segmentation. Recent work in the context of neuro-surgery also addressed the spine. Kaminsky [3] separated the spinal segments by manually delineating the joint parts of the vertebrae. Typical processing times were 1-2 hours. Kaus [4] and Leventon [5] used statistical shape models to segment individual vertebral bodies. Leventon used a trade-off between statistical models and Level Sets which dominate the final phase of the segmentation process. While theoretically sound current statistical models are not suited to cope with fine details of a given shape, in particular if curvature is high. This may be caused by insufficient sets of training data. The problem of the generation of suitable statistical shape models of the spine is discussed in Vrtovec [6].

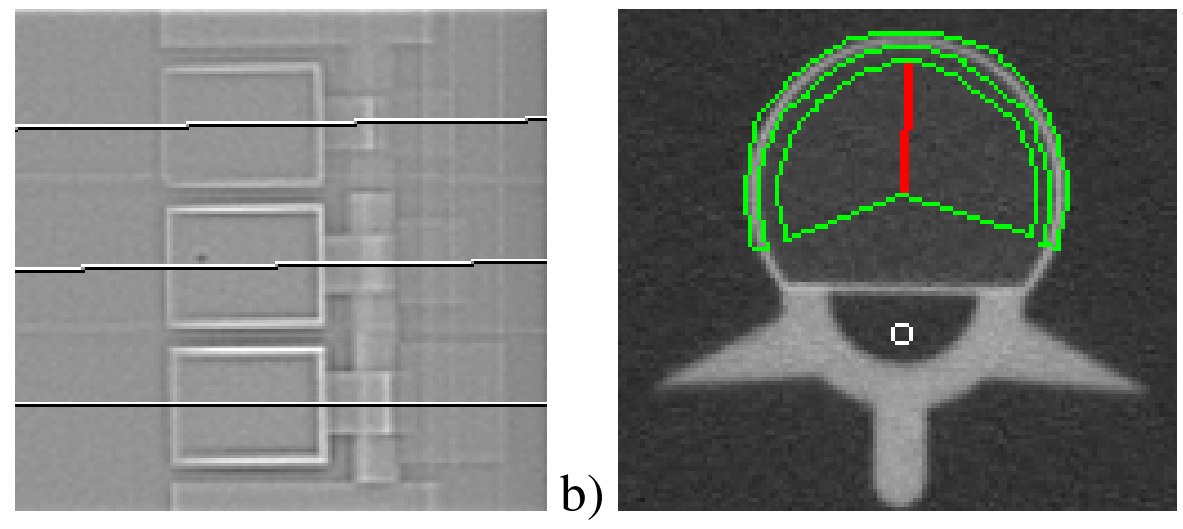

**Fig. 1.** State-of-the-Art QCT using a semi-anthropometric phantom of the lumbar spine, the European Spine Phantom (ESP): a) sagittal topogram with mid-vertebral slices, b) axial slice through L2 with segmented cortical and trabecular compartments to be quantitatively analyzed

**Previous and Current Work.** One of the coauthors (W. Kalender) developed a 2D segmentation and the quantitative BMD analysis of the vertebral bodies as a diagnostic tool for osteoporosis. This work has been integrated as the Osteo option on CT scanners from Siemens (Siemens Medical Solutions, Forchheim, Germany) (see Fig. 1). More recently we developed a true 3D segmentation procedure for CT images of the proximal femur that showed excellent precision [7]. It is a multi-step procedure combining volume growing using local adaptive thresholds with a variety of morphological operations to fill holes. It also integrates a semi-automatic step to separate the femur from the acetabulum. However, these methods can only partly be used in the spine because the anatomy is more complex. First, the separation of the anatomical parts is more complicated than in the hip and secondly, the cortex of the vertebral body is very thin.

In this work we developed a new combined approach, that proceeds from coarse to fine. Typically, individual bones have to be separated first to guide the rest of the segmentation [3, 8]. Prior to using the main ideas from [7] a deformable 3D balloon surface in a constrained search space is iteratively moved towards the outer cortical surface on radially emerging profiles. The reason for the initial use of the balloons is their strength to quickly segment the main shape of the vertebrae, bridging gaps in the partly very thin cortical walls and thus connect separated pieces of the cortex caused by noise and fractures.

## 2 Materials and Methods

### 2.1 Segmentation Procedure

For deformable models two formal representations exist: a) explicit deformable models using meshes [9] and b) implicit formulations also called Level Set methods [10]. To avoid the computational cost and numerical instability associated with the implicit gradient flow methods of type b) we have chosen an explicit three-dimensional triangle mesh.

**Constraints.** First the user is asked for the centers of the vertebrae. Constraints then define the search spaces for the subsequent steps. Their aim is the coarse separation of the vertebral bodies with a boolean combination of simple geometric shapes. Cylinders with angular capped ends enclose each vertebral body. Using the approximate marked centers of the vertebral bodies and the spinal channel, which is detected by a rolling ball procedure similar to Kaminsky [3], and the vertebral disks, which are automatically approximated by planes, the extents of the cylinders are defined. Additionally our implementation offers the possibility of user interaction should one of the constraint finding or segmentation steps fail.

**Coarse to Fine Segmentation.** The balloon model chosen here consists of a content-adaptive triangle mesh with automatic regularization. From a theoretical point of view the procedure solves the Euler-Lagrangian equations of motion. They are coupled to a minimization problem of internal and external energy of the balloon. The internal energy depends on the network connectivity forces whereas the external energy is derived from the gray values of the image. In our case the periosteal border of the cortical shell attracts the vertices of the balloon while the internal forces try to pertain the smooth and energy minimizing shape of a sphere. Due to the compromise between image gray values and surface model based forces the procedure is able to bridge gaps in the presence of noisy spurious contours and low contrast.

Practically the problem is solved by integrating the equations of motions:

$$m \cdot \ddot{\vec{p}} + \gamma \cdot \dot{\vec{p}} = \vec{f}_{ball} = \vec{f}_{img} + \vec{f}_{smg} + \vec{f}_{\text{inf}} \qquad (1)$$

for every vertex at position $\vec{p}$ using finite differences [9]. The force $\vec{f}_{ball}$ typically is the sum of the inflation forces $\vec{f}_{inf}$, of some external image forces $\vec{f}_{img}$ and of smoothing forces $\vec{f}_{smg}$ determining the viscosity of the balloon surface. In our implementation mass $m$ is set to 1, while damping $\gamma$ and inflation forces $\vec{f}_{inf}$ are ignored. As a consequence of the latter the 'leaking out problem' is prevented. If there are no attractive points close to the boundary of the balloon surface the points do not move. They are only influenced by the viscosity term, and their neighbors which trail them. Hence points not attracted by image forces are able to interpolate gaps in the contours.

**Local Adaptation.** The idea behind local vertex insertion is that in regions with high movement additional degrees of freedom are needed for the surface to adapt to fine anatomical structures. In the vertebral bodies this occurs when the balloon surface flows into the sharp corners at the rim of the endplates or the pedicles where the anatomy shows high curvature. To ease the balloon passing into these parts new degrees of freedom (new vertices) are inserted into the triangle mesh [11], which make the surface locally more flexible.

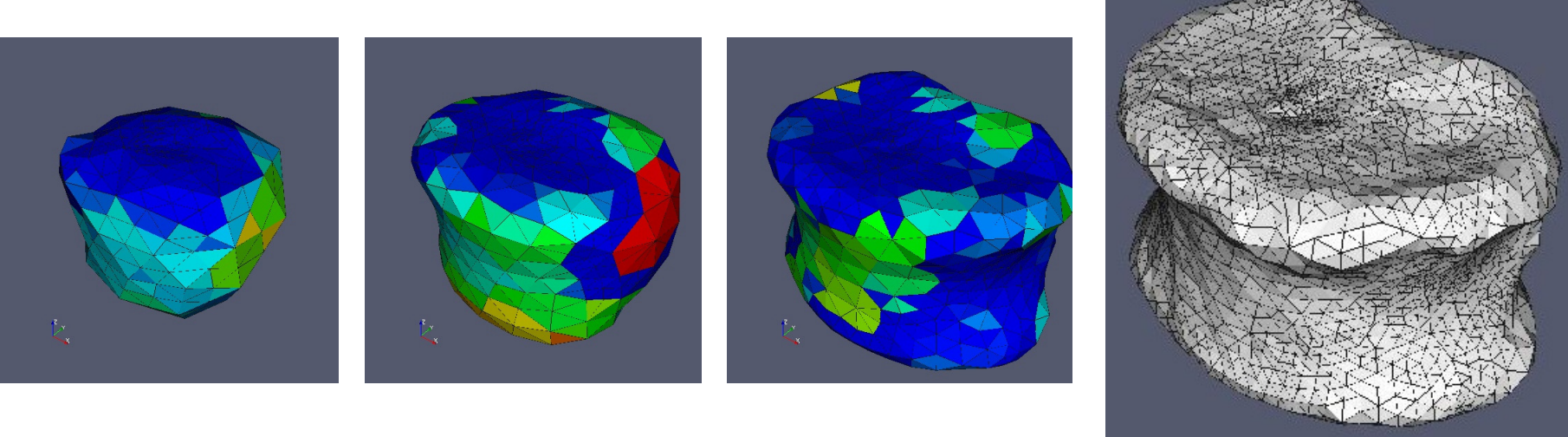

**Fig. 2.** Four selected iterations of the deformable balloon model inflated in an individual vertebra. Triangles, which are candidates for vertex insertion are coded in light colors. Right-handed is the final segmentation of the balloon stage

**Image and Smoothing Forces.** The three-dimensional segmentation problem is reduced to a one-dimensional problem for every surface vertex by iteratively moving points along profiles that radially emerge from the balloon surface [9]. For every vertex the gray value profile along the normal is sampled with sub-voxel resolution. Thus the profile analysis determines $\vec{f}_{img}$. However, due to the smoothness term $\vec{f}_{smg}$ that is a major regularizing part of the deformable model segmentation of locations with high curvature remains obtuse (see the rightmost picture in Fig. 2 and the left-hand image in Fig. 3). Smoothing forces are derived from the first order neighborhood of a vertex modeled as network of springs.

**Multi-seeded Volume Growing.** At this state where the strengths of the balloon have been exhausted, the border of the balloon is searched for points that pass the criteria developed by Kang [7]. A low and a high threshold is derived from the intersection of two Gaussian distributions fitted to the two peaks of the bimodal histogram of the gray values in the search space of one vertebra. Values below the lower and above the higher threshold represent soft and bone tissue, respectively. In the transition zone between the two thresholds, a local noise adaptive criterion is used to separate bone from soft tissue. Voxels satisfying the threshold criteria on the surface are used as seed points for a local volume growing process that is followed by a morphological closing and hole filling. During this phase parts of the processes are added as can be seen in the left-hand vertebral body in Fig. 3.

**Pedicle Cut.** In order to precisely determine the volume of the vertebral body the processes must be removed by an automatic procedure. Briefly, we try to find the thinnest part in the pedicles to cut-off the process. The methods used here are derived from the morphological concepts of ultimate erosion and Skeleton by Influence Zones (SKIZ) [12]. Influence zones are related to the concepts of Euclidean Distance Trans-form of regions and the Voronoi partition of space between them. First the segmentation result is eroded to its main components, the vertebral body residual and the residuals of the processes.

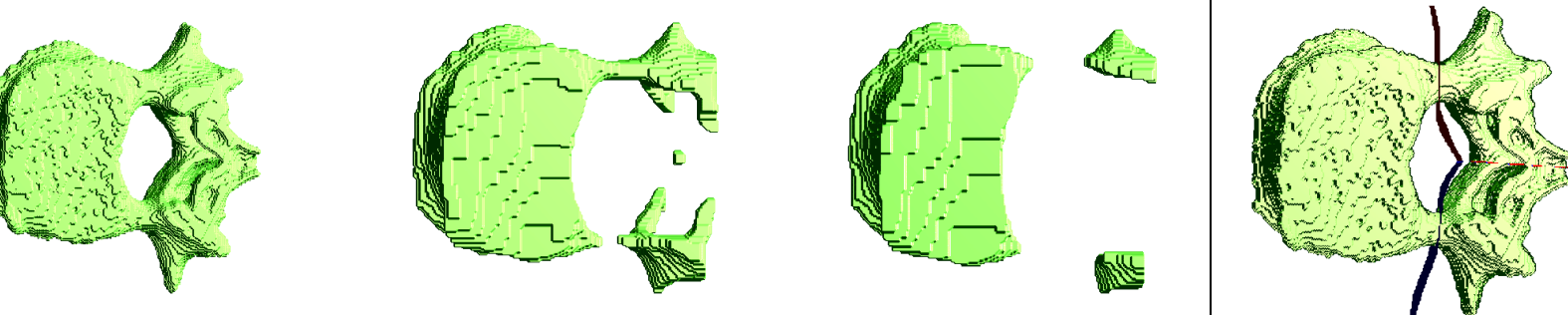

**Fig. 3.** Three selected iterations from the morphological cut off procedure and the final partition with the dissection surfaces that separate the processes from the vertebral body

The dissection surface is found by non-intersecting parallel dilation operations of the residuals of the vertebral body and the process. The contact areas of the dilated residuals yield the desired cutting surfaces (Fig. 3).

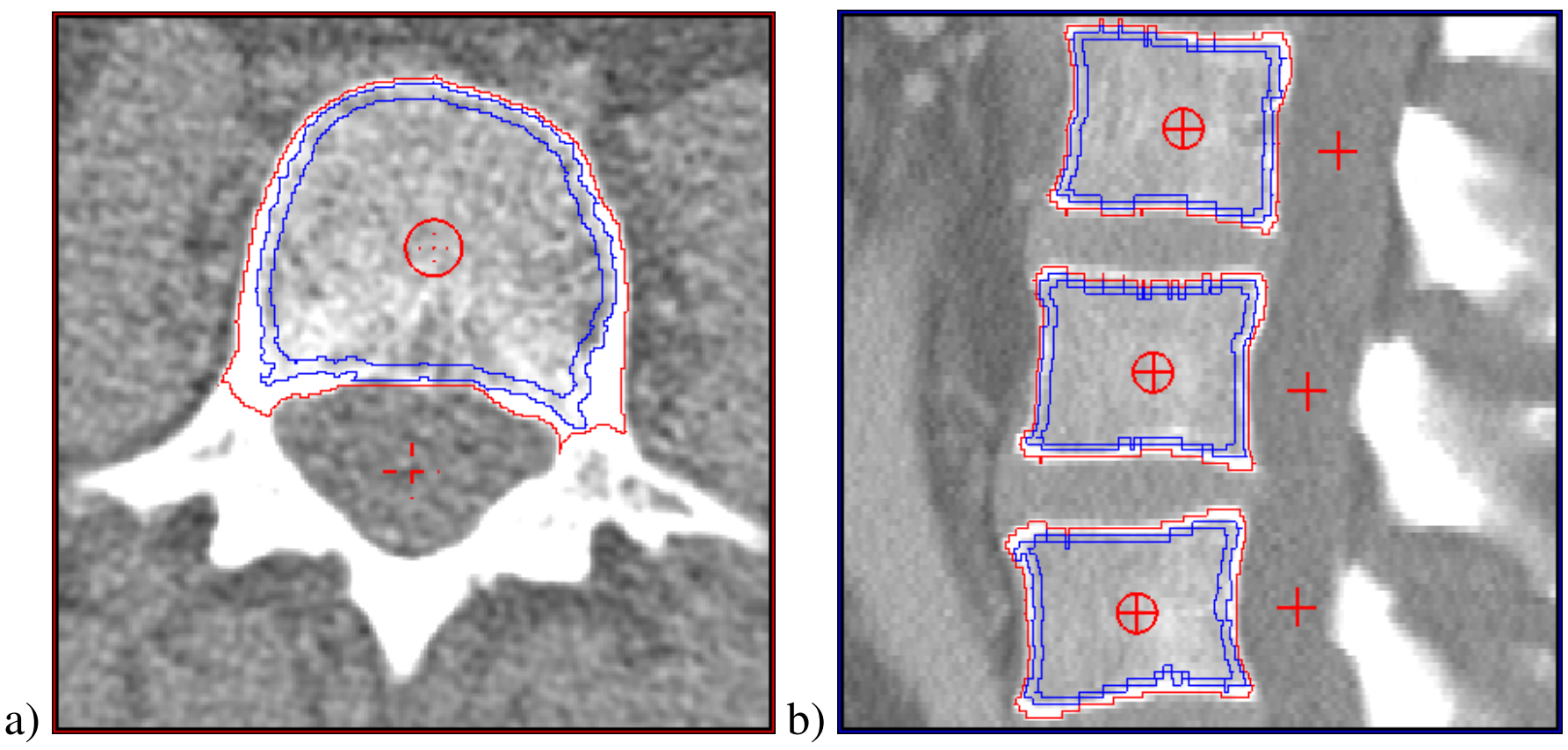

**Fig. 4.** Segmentation results from a patient dataset: a) transversal view, b) sagittal view. The markers denote the middle of the vertebral body and the center of the spinal channel

**Trabecular Compartment.** A gray value based locally adaptive erosion operation removes voxels with high gray values located in the cortical shell of the vertebral body. The high threshold described above is used for this purpose. Afterwards an optional homogeneous erosion operation is used to peel off the subcortical bone in order to assess the pure trabecular bone of the vertebral body (Fig. 4).

### 2.3 Performance

**Accuracy** was assessed using the European Spine Phantom, a geometrically defined, semi-anthropomorphic phantom [13]. CT datasets were obtained on a Siemens Sensation 16 at 120 kV and a slice thickness of 1 mm. In order to investigate the influence of noise the radiation dose was varied: three different time current products (580, 145 and 36 mAs) were used. Relative to 580 mAs the other two settings increase noise by factors of 2 and 4. Each dataset was analyzed three times by the same operator. BMD and volume were determined in the trabecular compartments. Results were averaged and compared to the nominal values of the phantom. For volume measurement the subcortical bone was included, for the BMD measurement it was excluded. By this evaluation the correctness of the segmentation is appraised. Even so it shall be noted explicitly that in this setup calibration for BMD and data acquisition is included into the quality feedback loop and also affect the error statistics.

**Precision** was analyzed using clinical routine abdominal scans from 10 patients. CT acquisition was again performed on a Siemens Sensation 16 (60 mAs, 120 kV, slice thickness 1 mm). For each patient datasets with three different fields of view (FOV) (150, 250, 350 mm) with corresponding in-plane pixel sizes of 0.3, 0.5 and 0.7 mm were used. Again all datasets were analyzed three times by the same operator. Intra-operator precision was determined for BMD and volume of the vertebral bodies L1-L3 as follows. For each patient and each of the three analyses results for L1-L3 were averaged. Then all 10 patients were averaged resulting in a percent coefficient of variation (%CV) per analysis. Finally root mean square averages were calculated for the three analyses resulting in an average %CV and a corresponding standard deviation SD. The purpose of this study is to evaluate how much the required little user-interaction influences the segmentation results.

## 3 Results

Fig. 4 and 5 illustrate the results of our multi-step segmentation procedure in a patient dataset and the ESP. The trabecular compartments are embedded in the outer segmentation.

**Accuracy.** Fig. 5 illustrates the noise dependence of segmentation results in the ESP (FOV 150 mm). The total VOI is shown as light outer and the trabecular VOIs as dark inner contours. The most inner VOI excludes the subcortical bone. Obviously this is less important in a well defined phantom than in patients.

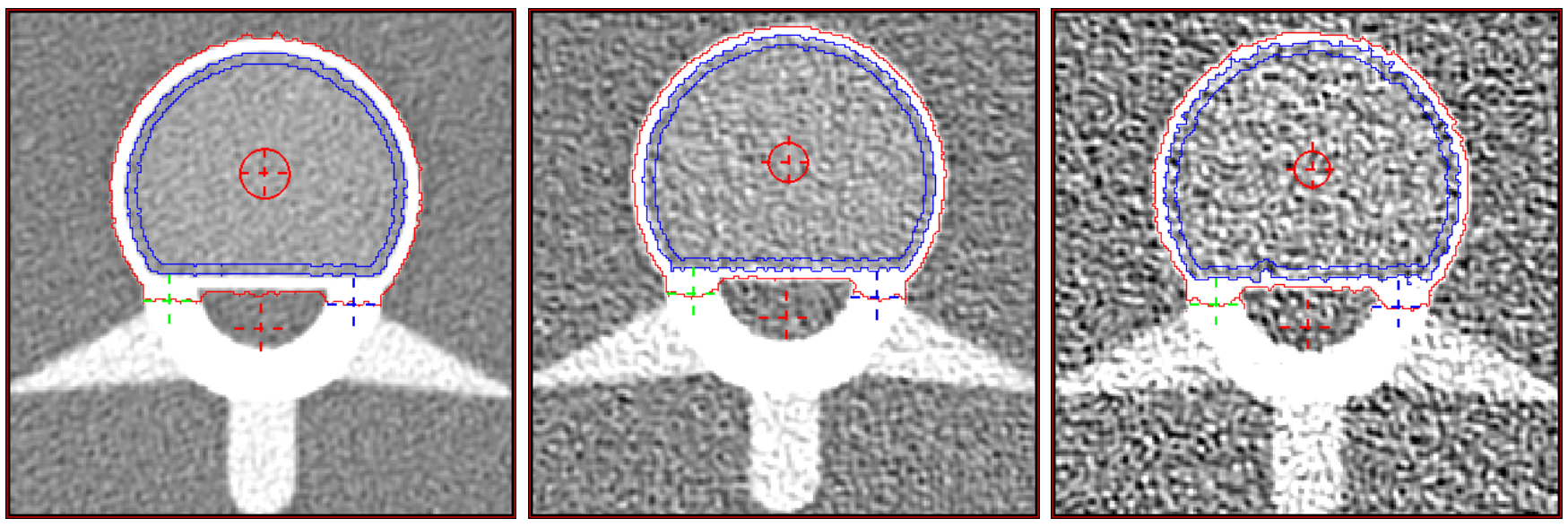

**Fig. 5.** Segmentation results of the ESP, vertebra L2, different noise levels obtained from three different exposures (580, 145, 36 mAs). Corresponding standard deviations measured in the 'soft tissue' region were approximately 50, 100 and 200 HU

For the purpose of this publication BMD measurements were only performed in the most inward VOI, but the outer trabecular VOI was used to determine volume. Tab. 1 summarizes the results. With respect to segmentation volume is the most relevant measurement because BMD of the most inner VOI is hardly affected by segmentation. As can be seen accuracy errors for volume are well below 4% which means that the segmentation errors are in the order of a falsely segmented slice of the vertebral body in z-direction (e.g. 5000 voxels with FOV 150). This may be partly attributed to partial volume effects at the endplates in some data acquisitions and motivates further examinations. Generally accuracy can be improved by reducing noise and the reconstructed FOV of the CT images although not all results are consistent with this claim.

**Table 1.** Accuracy results: % Errors from nominal values:

| FOV [mm] | | 150 | | | 250 | | | 350 | | |
|---|---|---|---|---|---|---|---|---|---|---|
| Time current product [mAs] | | 580 | 145 | 36 | 580 | 145 | 36 | 580 | 145 | 36 |
| L1 | BMD | 0.37 | 0.68 | 1.47 | 0.80 | 1.36 | 0.77 | 0.54 | 1.78 | 0.68 |
| | Vol. | 0.38 | 0.42 | 0.96 | 0.06 | 0.71 | 2.21 | 0.25 | 1.32 | 3.43 |
| L2 | BMD | 1.11 | 0.92 | 0.45 | 0.52 | 0.51 | 0.24 | 1.09 | 0.84 | 0.26 |
| | Vol. | 3.48 | 0.31 | 1.20 | 3.14 | 0.09 | 0.17 | 1.96 | 0.03 | 0.17 |
| L3 | BMD | 0.25 | 0.27 | 0.70 | 0.11 | 0.11 | 0.86 | 0.25 | 0.18 | 1.12 |
| | Vol. | 2.18 | 0.12 | 2.73 | 2.18 | 2.58 | 2.42 | 2.27 | 2.72 | 3.52 |

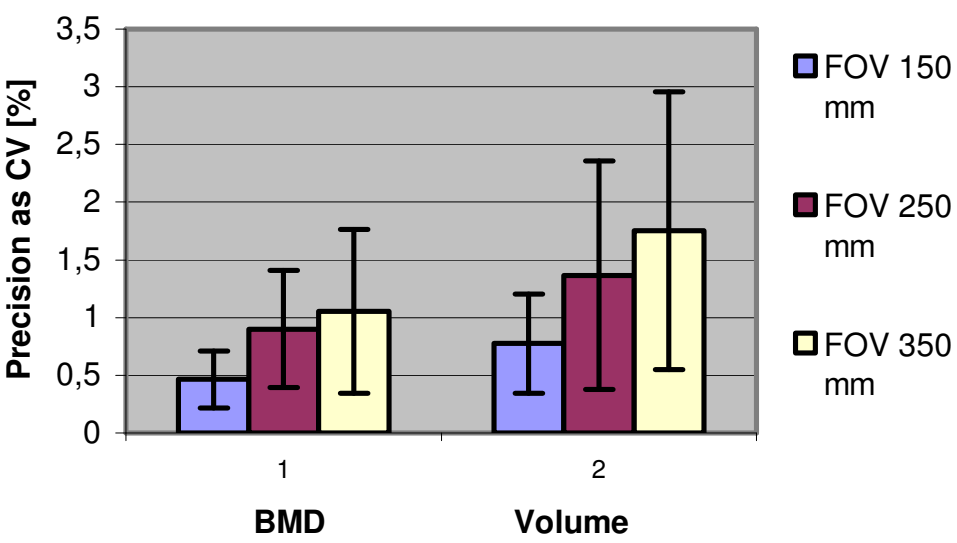

**Fig. 6.** Precision errors as coefficients of variation (RMS) and standard deviations

**Precision.** Results of the intra-operator analysis in patients are shown in Fig. 6. for the same parameters as in Tab. 1. Precision errors decreased for smaller FOVs. In patients the BMD results are stronger influenced by the segmentation because BMD even in the trabecular VOIs is not as homogeneous as in the ESP.

BMD errors of below 1% are very good results but it must be pointed out that here only the effect of the segmentation (no calibration for BMD) has been evaluated and that we still have to provide inter-operator precision and patient repositioning errors.

## 4 Conclusion

Our semi-automatic combined approach delivers very convincing segmentation results. It requires only a low degree of user interaction for initially placing points in the centers of the vertebral bodies and sometimes modifications of the constraints. Our procedure takes up to 15 minutes with a trained operator on a standard Pentium IV PC with 2.8 GHz and 1 GB of memory. Precision and accuracy results shown here are excellent but we will provide some more comprehensive evaluations. The performance characteristics indicate superiority of our 3D approach over current standard QCT analysis procedures in the field of osteoporosis. This would be an important step to improve fracture prediction and the therapeutic monitoring in this important bone disease.